%% file: main.tex
\documentclass[11pt]{article}

\usepackage[]{acl}

\usepackage{times}
\usepackage{latexsym}
\usepackage{todonotes}
\usepackage{booktabs}
\usepackage{url}
\usepackage{multirow}

\usepackage[T1]{fontenc}

\usepackage[utf8]{inputenc}

\usepackage{microtype}

\usepackage{inconsolata}

\usepackage{graphicx}

\title{Bangla Hate Speech Classification with Fine-tuned Transformer Models}

\author{Yalda Keivan Jafari \\
  Faculty of Computer Science,\\
  University of New Brunswick,\\ NB, Canada \\
  \texttt{yalda.keivan-jafari@unb.ca} \\\And
  Krishno Dey \\
  Faculty of Computer Science,\\
  University of New Brunswick, \\NB, Canada \\
  \texttt{krishno.dey@unb.ca} \\}

\begin{document}
\maketitle
\begin{abstract}
Hate speech recognition in low-resource languages remains a difficult problem due to insufficient datasets, orthographic heterogeneity, and linguistic variety. 
Bangla is spoken by more than 230 million people of Bangladesh and India (West Bengal).  Despite the growing need for automated moderation on social media platforms, Bangla is significantly underrepresented in computational resources.  
In this work, we study Subtask 1A and Subtask 1B of the BLP 2025 Shared Task on hate speech detection. We reproduce the official baselines (e.g., Majority, Random, Support Vector Machine) and also produce and consider Logistic Regression, Random Forest, and Decision Tree as baseline methods. We also utilized transformer-based models such as \textit{DistilBERT}, \textit{BanglaBERT}, \textit{m-BERT}, and \textit{XLM-RoBERTa} for hate speech classification.
All the transformer-based models outperformed baseline methods for the subtasks, except for \textit{DistilBERT}. Among the transformer-based models, \textit{BanglaBERT} produces the best performance for both subtasks. Despite being smaller in size, \textit{BanglaBERT} outperforms both \textit{m-BERT} and \textit{XLM-RoBERTa}, which suggests language-specific pre-training is very important.
Our results highlight the potential and need for pre-trained language models for the low-resource Bangla language.  
\end{abstract}

\input{sections/intro}
\input{sections/background}
\input{sections/method}

\input{sections/implementation}
\input{sections/evaluation}
\input{sections/challenges}
\input{sections/contributions}
\input{sections/conclusion}

\input{sections/limitations}
\input{sections/ethics}



\bibliography{references}

\appendix

\input{appendix/detailed-result}

\end{document}

%% file: sections/intro.tex
\section{Introduction}
\label{sec:intro}
The widespread usage of the internet has grown significantly over the past few decades, social media platforms have become much more popular worldwide. Improved access to the internet has allows people to express their thoughts and information online. The use of social media has been rising rapidly and does not appear to be slowing down\footnote{https://www.pewresearch.org/internet/fact-sheet/social-media/}. 
Social media has become common in public discussion, but it allows people to spread hatred. Rapid increase in toxic and harmful content (hate speech) raises substantial safety concerns in these online platforms. Hate speech can be defined as harmful, offensive, or derogatory language targeting individuals or groups \cite{ishmam2019hateful}. As a result, automatic detection of hate speech has become an important Natural Language Processing (NLP) task. There has been a lot of development in hate-speech detection in high-resource languages such as English, while developments in low-resource languages such as Bangla (Bengali) are still limited \cite{subramanian2023survey}. Such limitations is caused by the scarcity of high-quality annotated datasets and the complex linguistic properties of the language. 

The Bangla Language Processing (BLP) Workshop at IJCNLP–AACL 2025 introduced two shared tasks aimed at advancing NLP research for Bangla \cite{blp2025-overview-task1, hasan2025llm}. In this work, we consider Task 1, which focuses on hate speech classification across multiple dimensions, including hate type and target identification. This task is particularly relevant in the context of Bangla social media platforms, where hate speech frequently appears in conversations on politics, religion, and gender.

We aim to develop robust hate speech classifiers utilizing the dataset provided by the shared task organizers. The task includes three subtasks; we focus on Subtask 1A (Hate Type Classification) and Subtask 1B (Target Identification). Both tasks are multi-class classification problems involving short Bangla YouTube comments. We utilize fine-tuned transformer-based models (\textit{DistilBERT} \cite{sanh2019distilbert}, \textit{BanglaBERT} \cite{bhattacharjee-etal-2022-banglabert}, \textit{m-BERT} \cite{devlin2018pretraining}, and \textit{XLM-RoBERTa} \cite{xlm-roBERTa}) for classification. Then we compare the performance of transformer-based models with the official baseline.   

We highlight the contributions of our work in the following:

\begin{itemize}
    \item We reproduce official baselines (e.g., majority, random, and SVM).
    \item We also produce results for Logistic Regression (LR),  Random Forest (RF), and Decision Tree (DT), and consider them as baseline methods.
    \item We fine-tune several transformer-based models, including \textit{DistilBERT}, \textit{BanglaBERT}, \textit{m-BERT}, and \textit{XLM-RoBERTa}, and present a detailed evaluation.
    \item We offer a comprehensive discussion of modeling challenges and dataset limitations.
\end{itemize}

The remainder of this paper is organized as follows. Section \ref{sec:background} reviews prior work and provides background to hate-speech detection in Bangla. Section \ref{sec:method} describes the methodology followed for this project. Section \ref{sec:implementation} discusses implementation details. Section \ref{sec:evaluation} presents evaluation results. Section \ref{sec:challenges} outlines challenges. Section \ref{sec:contributions} discusses the individual contributions.  Section \ref{sec:conclusion} describe the conclusion. 

%% file: sections/background.tex
\section{Background}
\label{sec:background}
Social media has become an integral part of everyone's daily life. It enables quick communication, easy information sharing, and the exchange of opinions across geographical boundaries. However, the emergence of hate speech and harsh language on social media platforms has also been facilitated by this expression of freedom. Unfortunately, not much attention has been paid to identifying hate speech on Bangla social media. It lacks a good-quality dataset. The vocabulary used in social media often differs greatly from that used in traditional print media. As a result, it is challenging to automatically identify Bangla hate speech.

\subsection{Hate Speech Detection}
Traditionally, the identification of hate speech has been defined as a supervised classification problem \cite{mossie2020vulnerable, alkomah2022literature}. Early research relied on lexicon-based and rule-based heuristics, but these techniques frequently failed to identify implicit or context-specific hatred \cite{cavicchio2024rule, gitari2015lexicon, asghar2017sentence}. 
More recent research employs machine learning classifiers such as logistic regression and SVMs \cite{mossie2020vulnerable}. 
In order to detect online vulnerability or dangerous content, traditional supervised learning techniques have been used to extract linguistic, behavioral, and user-level data \cite{subramanian2023survey}. For instance, decision trees (DT), random forests (RF), and logistic regressions (LR) models have been utilized for classifying harmful or abusive texts on social media platforms \cite{kiilu2018using,  toktarova2023hate, abro2020automatic}. Similarly, compared to basic lexical features, neural feature-based classifiers have improved detection accuracy by capturing contextual patterns in conversational data \cite{badjatiya2017deep, putra2022hate}. To improve prediction performance in online safety detection tasks, other research has also integrated text content with user metadata (e.g., images and videos) into a multi-modal structure \cite{karim2022multimodal, rana2022emotion}.

With the emergence of transformer-based models, performance has greatly increased, and deep representations have made contextual comprehension possible.
Models such as \textit{BERT} and \textit{RoBERTa} have shown substantial improvements in hate speech detection tasks\cite{devlin2018pretraining, xlm-roBERTa}. These models are capable of capturing long-range dependencies and subtle linguistic cues in language.  Several works have employed transformer architectures for other safety-related (Natural Language Processing) NLP tasks \cite{chiu2021detecting, mutanga2020hate, alonso2020hate, ghosh2023transformer}. These developments demonstrate how transformer-based embeddings are useful for detecting dangerous or susceptible user behavior on online platforms.

\subsection{Hate Speech Detection in Low-Resource Languages}
Research and practical applications in the field of NLP are mostly centered around high-resource languages that typically have large annotated corpora, well-established tools and libraries, and robust language models trained on large data. Researchers seem to prioritize analyzing high-resource languages and overlooking other languages spoken by billions of people \cite{bender2019benderrule}. 

Research in low-resource languages faces several obstacles, such as data sparsity, morphological richness, and inconsistent orthography. Studies in low-languages (e.g., Bangla, Hindi, and Urdu) highlight these challenges and demonstrate the need for further developments. There has been some recent development in these languages with newly curated datasets and language-specific pre-trained transformers. However, there are still a lot of rooms for further improvement and development.  In this work, we only focus on Bangla language processing. 

Bangla is one of the most widely spoken languages in the world\footnote{https://www.ethnologue.com/insights/ethnologue200/}. Despite being the seventh most spoken language globally, Bangla remains particularly under-resourced. Such room for research has boosted the development of a good-quality dataset in recent times \cite{romim2021hs, romim2021hate}. Additionally, a number of standards for identifying hate speech have been established \cite{alam2020bangla, das2022hate, romim2021hate}. The creation of the pretrain transformer (BanglaBERT) has created new research prospects \cite{bhattacharjee-etal-2022-banglabert}.  

\subsection{Bangla NLP and Pre-trained Models}
The first large-scale Bangla-specific transformer model was presented by \textit{BanglaBERT} \cite{bhattacharjee-etal-2022-banglabert}, which was trained on about 27GB of Bangla text.
\textit{BanglaBERT} significantly improved performance across a wide range of Bangla NLP tasks by capturing language-specific syntactic and semantic patterns that multilingual models typically overlook \cite{ghosh2023transformer, alam2020bangla}. \textit{BanglishBERT} is another variant that is pre-trained on English and Bangla data \cite{bhattacharjee-etal-2022-banglabert}.  The term Banglish refers to the use of the English alphabet to write Bangla. 

On the other hand, multilingual transformer models such as \textit{XLM-RoBERTa} \cite{xlm-roBERTa} and \textit{mBERT} \cite{devlin2018pretraining} provide support for Bangla as part of their multilingual pretraining. However, due to their representational capacity being shared across numerous languages, their performance frequently lags behind monolingual Bangla models, especially for tasks demanding deeper syntactic knowledge \cite{alam2020bangla}. For high-precision Bangla NLP applications, monolingual models remain the best option.

%% file: sections/method.tex
\section{Methodology}
\label{sec:method}
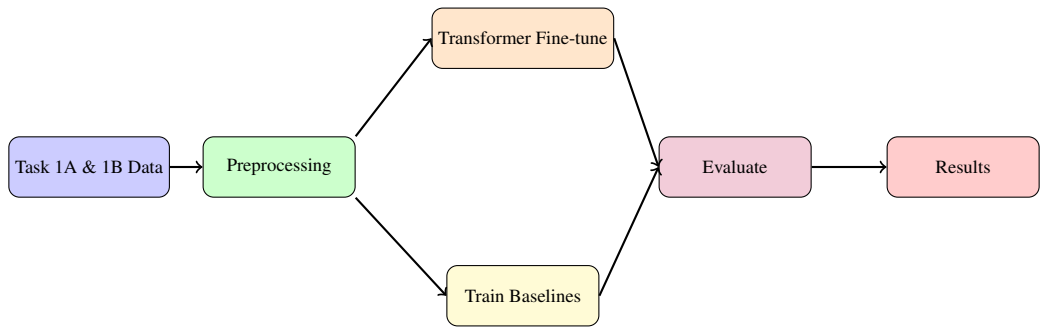
\begin{figure*}[]
    \centering
    \scriptsize
    \begin{tikzpicture}[node distance=1cm, auto]

        \node[draw, rectangle, rounded corners, minimum width=2cm, minimum height=.80cm, text centered, fill=blue!20] (step1) {Task 1A \& 1B Data};
        \node[draw, rectangle, rounded corners, minimum width=2cm, minimum height=0.80cm, text centered, fill=green!20, right of=step1, xshift=1.5cm] (step2) {Preprocessing};
        \node[draw, rectangle, rounded corners, minimum width=2cm, minimum height=0.80cm, text centered, fill=orange!20, above right of=step2, xshift=2.5cm, yshift=1cm] (step3a) {Transformer Fine-tune};
        \node[draw, rectangle, rounded corners, minimum width=2cm, minimum height=0.8cm, text centered, fill=yellow!20, below right of=step2, xshift=2.5cm, yshift=-1cm] (step3b) {Train Baselines};
        \node[draw, rectangle, rounded corners, minimum width=2cm, minimum height=0.8cm, text centered, fill=purple!20, right of=step2, xshift=5cm] (step4) {Evaluate};
        \node[draw, rectangle, rounded corners, minimum width=2cm, minimum height=.80cm, text centered, fill=red!20, right of=step4, xshift=2cm] (step5) {Results};

        \draw[->, thick] (step1) -- (step2);
        \draw[->, thick] (step2.north east) -- (step3a.west);
        \draw[->, thick] (step2.south east) -- (step3b.west);
        \draw[->, thick] (step3a.east) -- (step4.west);
        \draw[->, thick] (step3b.east) -- (step4.west);
        \draw[->, thick] (step4.east) -- (step5.west);

    \end{tikzpicture}
    \caption{Methodology Workflow for Bangla Hate Speech Classification}
    \label{fig:methodology_workflow}
\end{figure*}

This section describes our experimental methodology (see Figure \ref{fig:methodology_workflow}). We start with a brief overview of the dataset, then talk about our pre-processing steps, and present in-depth explanations of the models used in this study. 

\subsection{Dataset}
\label{tab:data-split}
\begin{table}[h!]
\centering
\begin{tabular}{lccc}
\hline
\textbf{Split} & \textbf{Original} & \textbf{Downsampled (1/3)} \\
\hline
Train & 35{,}522 & 11{,}840 \\
Dev   & 2{,}512  & 837 \\
Test  & 10{,}200 & 3{,}400 \\
\hline
\end{tabular}
\caption{Dataset sizes before and after downsampling to one-third.}
\end{table}

We use the BLP 2025 Shared Task dataset, which contains YouTube comments annotated for:
\begin{itemize}
    \item \textbf{Subtask 1A (Hate Type)}: Abusive, Sexism, Religious Hate, Political Hate, Profane, None
    \item \textbf{Subtask 1B (Hate Target)}: Individuals, Organizations, Communities, Society
\end{itemize}
Each instance contains an ID, the raw Bangla text, and a label. Comments include hateful expressions, informal spellings, and emojis. The dataset for all the tasks has the same number of samples.  Due to the computational complexity, we downsample the data by one-third.
Table \ref{tab:data-split} shows the total number of samples before and after downsampling for the train, dev, and test sets. 

\subsection{Preprocessing}\
We perform several pre-processing steps to prepare the BLP-2025 shared task 1 dataset for analysis and classification. We perform an extensive cleaning phase, during which special characters, URLs, emojis, and punctuation are eliminated. Tokenization is used to separate the text into individual words or tokens. Then we eliminate all of the stop words, which are generally low-content words that are used frequently in a language. The elimination of stop words enables the classification algorithm to concentrate on the keywords. These pre-processing steps further enhanced the quality of the dataset for subsequent analysis and classification tasks.


\subsection{Baselines}
The shared task uses the baselines (i.e., majority, random, and n-gram SVM) to evaluate the task performance. Along with these three baselines, we also consider LR, RF, and DT as baseline methods.  For comparison, we reproduce all six baselines commonly used in text classification tasks. These baselines serve as benchmarks for assessing the efficacy of our proposed (transformer-based) models.

We reproduce the following baselines:
\begin{itemize}
    \item \textbf{Majority}: Always predict the most frequent class.
    \item \textbf{Random}: Uniform random class selection.
    \item \textbf{n-gram SVM}: Support Vector Machine (SVM) classifiers using TF–IDF features.
    \item \textbf{Logistic Regression}: A multinomial logistic regression model trained on count vectorizer uni-gram features.
    \item \textbf{Random Forest}: An ensemble of learning method for classification with count vectorizer uni-gram features.
    \item \textbf{Decision Tree}: A single decision tree classifier with count vectorizer uni-gram features.
\end{itemize}

\subsection{Transformer-based Models}
We fine-tuned several transformer-based models, including \textit{DistilBERT}, \textbf{m-BERT}, \textit{XLM-RoBERTa base}, and \textit{BanglaBERT} for both the subtasks. Each model was trained for two epochs, a duration chosen for computational limitations. However, training for more epochs would result in better convergence on the test data; experimenting with more epochs is out of the scope of this project due to resource limitations. 
In order to enhance the model's performance, a batch size of 32 was utilized to accelerate the training procedure, and gradient accumulation was calculated every 32 data samples. The selection of a learning rate of 2e\textsuperscript{-5} was based on the principle that this rate facilitates more efficient learning of parameter estimates by the algorithm.



%% file: sections/implementation.tex
\section{Implementation Details}
\label{sec:implementation}

Our method combines traditional machine-learning baselines with transformer-based models.  We constructed baseline classifiers using scikit-learn and refined pre-trained models using the HuggingFace\footnote{\url{http://huggingface.co/}} ecosystem. All the implementation details are documented and provided here\footnote{\url{https://github.com/krishnodey/Hate_Speech_Classification}}.


\subsection{Libraries and Frameworks}
We install and use the following key libraries for developing and fine-tuning transformer models:

\begin{itemize}
    \item \texttt{transformers[torch]}: Helps in loading pre-trained models such as \textit{m-BERT} and \textit{BanglaBERT}, tokenization, training, and evaluation.
    
    \item \texttt{datasets}: Helps to load and preprocess training, validation, and test splits.
    
    \item \texttt{evaluate}: Helps to calculate evaluation metrics with standardized implementations, including macro F1-score, accuracy, precision, and recall.
    
\end{itemize}

We used the HuggingFace Trainer API to refine transformer models. The training parameters were as follows: batch size 16, maximum sequence length 128, learning rate $2e-5$, and optimizer AdamW. Each model employed its associated tokenizer, and the official dataset splits were used exactly as they were. We utilized the T4 GPU available for free of cost on Google Colab\footnote{\url{https://colab.research.google.com/}} for training and inferences. 

Traditional machine learning baselines, including majority baseline, random baseline, SVM, LR, RF, and DT models, are implemented using the \texttt{scikit-learn} toolkit. Text preprocessing and feature extraction use \texttt{pandas}, \texttt{NumPy}, \textit{n-gram}, \textit{TF–IDF}, and \textit{Count Vectorizer}.

\subsection{Project Workflow}
\begin{figure}[]
    \centering
    \scriptsize
    \begin{tikzpicture}[node distance=1.5cm, auto]
        \node[draw, rectangle, rounded corners, minimum width=3cm, minimum height=0.8cm, text centered, fill=blue!20] (data) {Data Loading and Preprocessing};
        \node[draw, rectangle, rounded corners, minimum width=3cm, minimum height=0.8cm, text centered, fill=green!20, below of=data] (baseline) {Training Classical Baseline Models};
        \node[draw, rectangle, rounded corners, minimum width=3cm, minimum height=0.8cm, text centered, fill=orange!20, below of=baseline] (transformer) {Fine-tuning Transformer Models};
        \node[draw, rectangle, rounded corners, minimum width=3cm, minimum height=0.8cm, text centered, fill=purple!20, below of=transformer] (evaluation) {Evaluating Models and Comparing Results};
        
        \draw[->, thick] (data) -- (baseline);
        \draw[->, thick] (baseline) -- (transformer);
        \draw[->, thick] (transformer) -- (evaluation);
    \end{tikzpicture}
    \caption{Workflow of the Bangla Hate Speech Classification Project}
    \label{fig:workflow}
\end{figure}
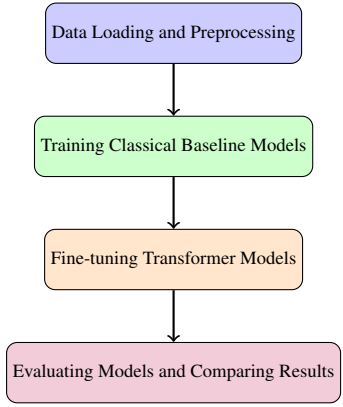

Figure \ref{fig:workflow} shows the workflow of this project. 

%% file: sections/evaluation.tex
\section{Evaluation}
\label{sec:evaluation}
The experimental setup, assessment measures, and outcomes of our Bangla hate speech categorization models are discussed in this section.

\subsection{Experimental Setup}
We utilized the official BLP 2025 dataset provided for Subtasks 1A and 1B. The dataset was divided into training, development, and test splits as provided.

For \textbf{transformer models}, we fine-tuned pre-trained models such as \textit{DistilBERT},  \textit{BanglaBERT}, \textit{m-BERT}, and \textbf{XLM-RoBERTa} using the HuggingFace \texttt{Trainer} API. Training details are in the following:

\begin{itemize}
    \item Optimizer: AdamW
    \item Learning rate: $2 \times 10^{-5}$
    \item Batch size: 16
    \item Maximum sequence length: 128 tokens
    \item Number of epochs: 2
\end{itemize}

Using scikit-learn, we trained Majority, Random, n-gram SVM, Logistic Regression, Random Forest, and Decision Tree classifiers for \textbf{baseline models}. We extracted features using TF-IDF and count-vector representations. While baseline models were trained on CPU, all experiments were conducted on a GPU-enabled environment (Google Colab T4) for transformer models.

\subsection{Evaluation Metrics}
We use the official evaluation metric \textbf{micro-averaged F1-score} as the primary evaluation metric. 
Additionally, we also calculate \textbf{accuracy}, \textbf{precision}, and \textbf{recall} for fair comparison.

\subsection{Experimental Results}



\begin{table}[]
    \centering
    \scriptsize
    \begin{tabular}{lcccc}
        \toprule
        \textbf{Model} & \textbf{Accuracy} & \textbf{Precision} & \textbf{Recall} & \textbf{Micro-F1} \\
        \midrule
        \multicolumn{5}{c}{Baselines}\\
        \midrule
        Random & 0.16 & 0.38 & 0.16 & 0.16 \\
        Majority & 0.57 & 0.33 & 0.57 & 0.57 \\
        SVM & 0.61 & 0.58 & 0.61 & 0.61 \\
        LR & 0.60 & 0.56 & 0.60 & 0.60 \\
        RF & 0.58 & 0.53 & 0.58 & 0.58 \\
        DT & 0.48 & 0.48 & 0.48 & 0.48 \\
        \midrule
        \multicolumn{5}{c}{Transformer-based Models}\\
        \midrule
        DistilBERT & 0.62 & 0.56 & 0.62 & 0.57 \\
        mBERT & 0.68 & 0.66	& 0.68 & 0.66 \\
        XLM-RoBERTa & 0.68 & 0.66	& 0.68	& 0.67 \\
        BanglaBERT & \textbf{0.71} &	\textbf{0.70} & \textbf{0.71} & \textbf{0.70} \\
        \bottomrule
    \end{tabular}
    \caption{Performance of Transformer-based and Baseline Models on BLP 2025 Subtasks 1A.  Bold numbers indicate the best scores among all the models. }
    \label{tab:results-1a}
\end{table}

In our study, we evaluate a wide range of models, such as \textit{m-BERT cased}, \textit{XLM-RoBERTa base}, \textit{BanglaBERT}, \textit{DistilBERT cased}, and the baseline models for classifying hate-speech on the BLP-2025 dataset. Table \ref{tab:results-1a} shows the classification report for subtask 1A. BanglaBERT achieved the highest accuracy of 0.71, surpassing all other models in our evaluation. BanglaBERT outperform other models with a precision and recall of 0.70 and 0.71, respectively. The official scoring metric for the shared task 1a was the micro F1 score. BanglaBERT achieved a micro-F1 score of 0.71 in our study. BERT multilingual base (m-BERT) achieved an F1 score of 0.66, while the XLM-RoBERTa base achieved an F1 score of 0.67, both surpassing the baselines. DistilBERT achieved an F1 score of 0.57, which is lower than the baseline methods SVM and LR. The micro F1 score of BanglaBERT outperforms other models in our study for this specific dataset of shared task 1. Among the baseline models SVM produces the best result with F1 score of 0.61, while the random baseline has the worst result with F1 score of 0.16.

Table \ref{tab:results-1b} illustrates the classification result for subtask 1B. Similar to subtask 1A, \textit{BanglaBERT} outperforms other transformer-based models and baselines. It achieves the micro F1 score of 0.68, an accuracy of 0.71, a precision of 0.69, and a recall of 0.71. \textit{XLM-RoBERTa} and \textit{m-BERT} also surpasses the baselines with F1 score of 0.66 and 0.67 respectively, while \textit{DistilBERT} fell short with the F1 score of 0.57. Among the baseline models SVM and LR models generates the best performance with F1 scores of 0.63 and 0.62, respectively.

One of the most critical findings from our analysis was the unexpected performance of the smaller model  \textit{BanglaBERT}, which outperformed larger models such \textit{m-BERT}, \textit{XLM-RoBERTa base}. This unexpected outcome highlights the importance of model architecture and the flexibility with which it can be adapted to the specifics of the dataset. Despite its smaller size, \textit{BanglaBERT} outperformed other models, suggesting its ability to capture the subtleties of language and context related to hate speech within the Bangla language. However, this superior performance can be attributed to its training on a Bangla dataset, enabling it to excel in this specific linguistic and contextual domain. This finding emphasizes the significance of pre-training data and architecture, in addition to size, when choosing models for particular NLP tasks.

The performance of SVM and LR is the best among the baseline methods. In subtask 1 A \textit{BanglaBERT} outperform SVM and LR by 0.9 and 1.0 terms of F1 score. Similarly,  in subtask 1B \textit{BanglaBERT} outperforms SVM and LR by 0.5 and 0.6 in terms of F1 score. For other models such as \textit{DistilBERT}, \textit{m-BERT}, and \textit{XLM-RoBERTa} these differences are substantially lower. All the transformer-based model used in this study produces better results than baseline methods. However, transformer-based models increases computational complexity. SVN and LR may still be a wonderful option in systems with minimal processing power and resources, where performance can be sacrificed. Table \ref{tab:classwise-results-1a} adn \ref{tab:classwise-results-1b} in the appexdix provides class-wise classification result for all the transformer-based models.

\begin{table}[]
    \centering
    \scriptsize
    \begin{tabular}{lcccc}
        \toprule
        \textbf{Model} & \textbf{Accuracy} & \textbf{Precision} & \textbf{Recall} & \textbf{Micro-F1} \\
        \midrule
        \multicolumn{5}{c}{Baselines}\\
        \midrule
        Random & 0.21 & 0.40 & 0.21 & 0.21 \\
        Majority & 0.60 & 0.36 & 0.60 & 0.60 \\
        SVM & 0.62 & 0.53 & 0.63 & 0.63\\
        LR & 0.62 & 0.54 & 0.62 & 0.62 \\
        RF & 0.60 & 0.52 & 0.60 & 0.60 \\
        DT & 0.48 & 0.48 & 0.48 & 0.48 \\
        \midrule
        \multicolumn{5}{c}{Transformer-based Models}\\
        \midrule
        DistilBERT & 0.65 & 0.59	& 0.65	& 0.57 \\
        mBERT & 0.69 & 0.66	& 0.69	& 0.66 \\
        XLM-RoBERTa & 0.69 & 0.66 & 0.69 & 0.66 \\
        BanglaBERT &\textbf{0.71} & \textbf{0.69}	& \textbf{0.71} & \textbf{0.68} \\
        \bottomrule
    \end{tabular}
    \caption{Performance of Transformer-based and Baseline Models on BLP 2025 Subtasks 1B.  Bold numbers indicate the best scores among all the models. }
    \label{tab:results-1b}
\end{table}








%% file: sections/challenges.tex
\section{Challenges}
\label{sec:challenges}

\subsection{Noisy Social Media Text}
Bangla YouTube comments often include non-standard spellings, emoticons, English code-mixing, and dialectal variances. The presence of such variants often introduces noise and increases linguistic ambiguity. Noisy text data are both difficult to preprocess and classify.

\subsection{Overlapping Hate Categories}
Many YouTube comments exhibit various types of hate simultaneously. However, the dataset contains only single-label annotations, which forces the model to choose one category even when several may apply. This discrepancy makes it difficult for models to learn about data, which may result in incorrect classification.


\subsection{Limited Context}
YouTube comments frequently appear as replies within longer discussion threads. The dataset does not include conversational context. The model must learn and classify each comment in isolation. This absence of information affects the ability to appropriately comprehend subtle or indirect hate sentiments.

%% file: sections/contributions.tex
\section{Individual Contributions}
\label{sec:contributions}
This project was carried out collaboratively by Krishno Dey and Yalda Keivan Jafari, with both members contributing equally to the overall development, experimentation, and documentation of the system. Individual responsibilities are detailed below.

\subsection{Contributions of Krishno Dey}

Krishno led the experimental workflow, implementation, and baseline reproduction. His key contributions include:

        \begin{itemize}
            \item Dataset Preparation and Preprocessing: Implemented preprocessing steps, including text normalization, cleaning, tokenization, and Data splitting for both subtasks.
            \item Baselines: Reproduced the official baseline models which serve as benchmarks for evaluating transformer models.
            \item Transformer Fine-tuning: Conducted fine-tuning experiments with BanglaBERT and mBERT, including configuration of hyperparameters, training procedures.
            \item Evaluation: Implemented the evaluation pipeline using micro-F1.
            \item Technical Writing: Drafted the Methodology, Implementation, and Evaluation sections.
        \end{itemize}

\subsection{Contributions of Yalda Keivan Jafari}

Yalda led the conceptual framing, multilingual model experiments, and literature integration. Her main contributions include:

        \begin{itemize}
            \item Background and Related Work: Conducted an in-depth review of prior work on hate speech detection and Bangla NLP resources, and authored the Background section.
            \item Transformer Model Experiments: Conducted fine-tuning experiments on XLM-RoBERTa, including configuration, training, and comparative analysis with monolingual models.
            \item Evaluation results and Discussion: Helped explain performance differences across models and contributed to the Challenges, Limitations, and Conclusion sections.
            \item Scientific Writing and Editing: Contributed to editing the full report for structure, flow, and academic writing quality.
            \item 
        \end{itemize}

\subsection{Joint Contributions}
Both members collaborated on the following aspects:
    \begin{itemize}
        \item Creating figures and tables.
        \item Conducting regular discussions to coordinate experiments and results.
    \end{itemize}

%% file: sections/conclusion.tex
\section{Conclusion}
\label{sec:conclusion}

We investigated Bangla hate speech detection across two subtasks (1A and 1B) of the BLP 2025 shared task. 
Our experiments show that all the transformer-based models (e.g., \textit{DistilBERT}, \textit{BanglaBERT}, \textit{m-BERT}, and \textit{XLM-RoBERTa}) outperform baseline methods except for \textit{DistilBERT} for both the subtasks. Among transformer-based models, \textit{BanglaBERT} produces the best performance. Pre-training on Bangla data helps the models to generalize well on both subtasks. 
Our study highlights the potential of transformer models for Bangla NLP and identifies areas requiring further exploration, such as handling noisy text and addressing overlapping hate categories.

%% file: sections/limitations.tex
\section*{Limitations}
\label{sec:limiations}






Our work has several limitations. 
\textbf{First}, the dataset provided for the shared task is downsampled to one-third to reduce computational complexity. Dwonsampling may restrict the generalization ability of the models.
\textbf{Second}, the comments in the dataset do not contain any contextual or conversational metadata. Such an absence of context may limit the ability of models to learn and generalize.
\textbf{Third}, Preprocessing may not generalize well across different Bangla dialects (e.g., Sylheti, Chittagonian) or informal writing styles. 
\textbf{Fourth}, larger transformer models such as \textit{XLM-RoBERTa Large}, \textit{BanglaBERT} Large, were not used due to computational limitations. Absence of those larger models may increase threats to the validity of our results discussion.
\textbf{Fifth}, although BanlgaBERT produced the best result among all the models, for subtask 1A it could not recognize \textbf{Profane} and \textbf{Abusive} classes (see Table \ref{tab:classwise-results-1a}). Similarly, DistilBERT also could not identify those two classes. Moreover, m-BERT and XLM-RoBERTa could not identify Abusive class for subtask 1A. 
\textbf{Finally}, we use the micro F1-score as the main evaluation metric, which helps lessen the influence of unequal label distributions, even though we did not specifically address class imbalance during training.

%% file: sections/ethics.tex
\section*{Ethics statement}
\label{sec:ethics}




This study focuses on automatic hate speech detection in Bangla social media text, a challenge with substantial ethical considerations. Any automated system created for this purpose runs the risk of being misinterpreted, misclassified, or abused. Classification of hate speech includes delicate sociocultural assessments.

Our study relies on pre-trained language models and annotated datasets,  both of which can contain existing societal biases. Subjectivity among annotators may result in inconsistent or culturally biased labeling, and during fine-tuning, transformer models may unintentionally magnify these biases. Furthermore, the downsampled dataset restricts representativeness, which could lead to inconsistent performance among groups, dialects, or writing styles.

Automated moderation systems generated from this research could have greater social consequences. Incorrect or overconfident model predictions may disproportionately damage marginalized groups, limit legal expression, or enable large-scale censorship. In contrast, under-detection could let dangerous information continue. Both outcomes illustrate the need for thorough and open deployment processes.

To mitigate these risks, we recommend the following:

\begin{itemize}
    \item \textbf{Human-in-the-loop moderation} to ensure that automated predictions support, rather than replace, human judgment.
    \item \textbf{Transparent reporting} of model limitations, intended use-cases, and performance disparities.
    \item \textbf{Fairness analysis} across demographic and linguistic subgroups to identify bias-related harms.
    \item \textbf{Restricted deployment} of automated systems in high-stakes or punitive contexts without rigorous auditing.
\end{itemize}

The models trained in this work are only meant for use in scholarly investigations. Without thorough analysis, contextual knowledge, and ethical supervision, they shouldn't be applied in real-world moderation or decision-making systems.

%% file: appendix/detailed-result.tex
\section{Detailed Classfication Result}
\label{sec:appendix}

Table \ref{tab:classwise-results-1a} and \ref{tab:classwise-results-1b} presents the detailed class-wise result for subtask 1A and 1B respectively.

\begin{table*}[t]
\centering
\scriptsize
\begin{tabular}{l l c c c}
\toprule
\textbf{Model} & \textbf{Class} & \textbf{Precision} & \textbf{Recall} & \textbf{F1-score} \\
\midrule

\multirow{6}{*}{DistilBERT}
 & None           & 0.37 & 0.18 & 0.24 \\
 & Religious Hate & 0.66 & 0.91 & 0.77 \\
 & Sexism         & 0.52 & 0.31 & 0.39 \\
 & Political Hate & 0.60 & 0.36 & 0.45 \\
 & Profane        & 0.00 & 0.00 & 0.00 \\
 & Abusive        & 0.00 & 0.00 & 0.00 \\
\midrule

\multirow{6}{*}{mBERT}
 & None           & 0.46 & 0.36 & 0.40 \\
 & Religious Hate & 0.75 & 0.85 & 0.79 \\
 & Sexism         & 0.59 & 0.50 & 0.54 \\
 & Political Hate & 0.73 & 0.68 & 0.71 \\
 & Profane        & 0.50 & 0.37 & 0.42 \\
 & Abusive        & 0.00 & 0.00 & 0.00 \\
\midrule

\multirow{6}{*}{XLM-RoBERTa}
 & None           & 0.47 & 0.39 & 0.42 \\
 & Religious Hate & 0.76 & 0.87 & 0.81 \\
 & Sexism         & 0.59 & 0.51 & 0.55 \\
 & Political Hate & 0.70 & 0.58 & 0.64 \\
 & Profane        & 0.53 & 0.24 & 0.33 \\
 & Abusive        & 0.00 & 0.00 & 0.00 \\
\midrule

\multirow{6}{*}{BanglaBERT}
 & None           & 0.50 & 0.56 & 0.53 \\
 & Religious Hate & 0.83 & 0.83 & 0.83 \\
 & Sexism         & 0.57 & 0.57 & 0.57 \\
 & Political Hate & 0.70 & 0.68 & 0.69 \\
 & Profane        & 0.00 & 0.00 & 0.00 \\
 & Abusive        & 0.00 & 0.00 & 0.00 \\
\bottomrule
\end{tabular}
\caption{Class-wise performance of transformer-based models on subtask 1A.}
\label{tab:classwise-results-1a}
\end{table*}

\begin{table*}[t]
\centering
\scriptsize
\begin{tabular}{llcccc}
\toprule
\textbf{Model} & \textbf{Class} & \textbf{Precision} & \textbf{Recall} & \textbf{F1-score}  \\
\midrule

\multirow{5}{*}{DistilBERT}
& None         & 0.57 & 0.02 & 0.03\\
& Society      & 0.56 & 0.19 & 0.29 \\
& Organization & 0.66 & 0.94 & 0.78 \\
& Community    & 0.58 & 0.46 & 0.52 \\
& Individual   & 0.00 & 0.00 & 0.00 \\
\midrule

\multirow{5}{*}{mBERT}
& None         & 0.45 & 0.25 & 0.32 \\
& Society      & 0.59 & 0.43 & 0.50\\
& Organization & 0.74 & 0.89 & 0.81\\
& Community    & 0.58 & 0.57 & 0.57\\
& Individual   & 0.40 & 0.11 & 0.17\\
\midrule

\multirow{5}{*}{XLM-RoBERTa}
& None         & 0.40 & 0.22 & 0.28\\
& Society      & 0.55 & 0.46 & 0.50\\
& Organization & 0.75 & 0.87 & 0.81\\
& Community    & 0.59 & 0.59 & 0.59\\
& Individual   & 0.41 & 0.17 & 0.24\\
\midrule

\multirow{5}{*}{BanglaBERT}
& None         & 0.44 & 0.17 & 0.24\\
& Society      & 0.59 & 0.58 & 0.58\\
& Organization & 0.79 & 0.89 & 0.83\\
& Community    & 0.51 & 0.64 & 0.57\\
& Individual   & 0.57 & 0.02 & 0.04\\
\bottomrule
\end{tabular}
\caption{Class-wise performance of transformer-based models on subtask 1B.}
\label{tab:classwise-results-1b}
\end{table*}

%% file: references.bib
@article{hasan2025llm,
      title={{LLM-Based Multi-Task Bangla Hate Speech Detection: Type, Severity, and Target}}, 
      author={Hasan, Md Arid and Alam, Firoj and Hossain, Md Fahad and Naseem, Usman and Ahmed, Syed Ishtiaque},
      year={2025},
      journal={arXiv preprint arXiv:2510.01995},
      url={https://arxiv.org/abs/2510.01995},
}

@inproceedings{blp2025-overview-task1,
    title = {{"Overview of BLP 2025 Task 1: Bangla Hate Speech Identification"}},
    author = "Hasan, Md Arid and Alam, Firoj and Hossain, Md Fahad and Naseem, Usman and Ahmed, Syed Ishtiaque",
    booktitle = "Proceedings of the Second International Workshop on Bangla Language Processing (BLP-2025)",
    editor = {Alam, Firoj
          and Kar, Sudipta
          and Chowdhury, Shammur Absar
          and Hassan, Naeemul
          and Prince, Enamul Hoque
          and Tasnim, Mohiuddin
          and Rony, Md Rashad Al Hasan
          and Rahman, Md Tahmid Rahman
    },
    month = dec,
    year = "2025",
    address = "India",
    publisher = "Association for Computational Linguistics",
}

@inproceedings{bhattacharjee-etal-2022-banglabert,
    title     = {{BanglaBERT: Lagnuage Model Pretraining and Benchmarks for Low-Resource Language Understanding Evaluation in Bangla}},
    author = "Bhattacharjee, Abhik  and
      Hasan, Tahmid  and
      Mubasshir, Kazi  and
      Islam, Md. Saiful  and
      Uddin, Wasi Ahmad  and
      Iqbal, Anindya  and
      Rahman, M. Sohel  and
      Shahriyar, Rifat",
      booktitle = "Findings of the North American Chapter of the Association for Computational Linguistics: NAACL 2022",
      month = july,
    year      = {2022},
    url       = {https://arxiv.org/abs/2101.00204},
    eprinttype = {arXiv},
    eprint    = {2101.00204}
}

@misc{devlin2018pretraining,
  author = {Devlin, Jacob and Chang, Ming-Wei and Lee, Kenton and Toutanova, Kristina},
  description = {[1810.04805] BERT: Pre-training of Deep Bidirectional Transformers for Language Understanding},
  note = {cite arxiv:1810.04805Comment: 13 pages},
  title = {{BERT: Pre-training of Deep Bidirectional Transformers for Language
  Understanding}},
  url = {http://arxiv.org/abs/1810.04805},
  year = 2018
}

@article{xlm-roBERTa,
  author       = {Alexis Conneau and
                  Kartikay Khandelwal and
                  Naman Goyal and
                  Vishrav Chaudhary and
                  Guillaume Wenzek and
                  Francisco Guzm{\'{a}}n and
                  Edouard Grave and
                  Myle Ott and
                  Luke Zettlemoyer and
                  Veselin Stoyanov},
  title        = {{Unsupervised Cross-lingual Representation Learning at Scale}},
  journal      = {CoRR},
  volume       = {abs/1911.02116},
  year         = {2019},
  url          = {http://arxiv.org/abs/1911.02116},
  eprinttype    = {arXiv},
  eprint       = {1911.02116},
  bibsource    = {dblp computer science bibliography, https://dblp.org}
}

@article{mossie2020vulnerable,
  title={Vulnerable community identification using hate speech detection on social media},
  author={Mossie, Zewdie and Wang, Jenq-Haur},
  journal={Information Processing \& Management},
  volume={57},
  number={3},
  pages={102087},
  year={2020},
  publisher={Elsevier}
}

@article{kiilu2018using,
  title={Using Na{\"\i}ve Bayes algorithm in detection of hate tweets},
  author={Kiilu, Kelvin Kiema and Okeyo, George and Rimiru, Richard and Ogada, Kennedy},
  journal={International Journal of Scientific and Research Publications},
  volume={8},
  number={3},
  pages={99--107},
  year={2018}
}

@article{chiu2021detecting,
  title={Detecting hate speech with gpt-3},
  author={Chiu, Ke-Li and Collins, Annie and Alexander, Rohan},
  journal={arXiv preprint arXiv:2103.12407},
  year={2021}
}

@article{abro2020automatic,
  title={Automatic hate speech detection using machine learning: A comparative study},
  author={Abro, Sindhu and Shaikh, Sarang and Khand, Zahid Hussain and Zafar, Ali and Khan, Sajid and Mujtaba, Ghulam},
  journal={International Journal of Advanced Computer Science and Applications},
  volume={11},
  number={8},
  year={2020},
  publisher={Science and Information (SAI) Organization Limited}
}

@inproceedings{badjatiya2017deep,
  title={Deep learning for hate speech detection in tweets},
  author={Badjatiya, Pinkesh and Gupta, Shashank and Gupta, Manish and Varma, Vasudeva},
  booktitle={Proceedings of the 26th international conference on World Wide Web companion},
  pages={759--760},
  year={2017}
}

@article{toktarova2023hate,
  title={Hate speech detection in social networks using machine learning and deep learning methods},
  author={Toktarova, Aigerim and Syrlybay, Dariga and Myrzakhmetova, Bayan and Anuarbekova, Gulzat and Rakhimbayeva, Gulbarshin and Zhylanbaeva, Balkiya and Suieuova, Nabat and Kerimbekov, Mukhtar},
  journal={International Journal of Advanced Computer Science and Applications},
  volume={14},
  number={5},
  year={2023},
  publisher={Science and Information (SAI) Organization Limited}
}

@article{subramanian2023survey,
  title={A survey on hate speech detection and sentiment analysis using machine learning and deep learning models},
  author={Subramanian, Malliga and Sathiskumar, Veerappampalayam Easwaramoorthy and Deepalakshmi, G and Cho, Jaehyuk and Manikandan, G},
  journal={Alexandria Engineering Journal},
  volume={80},
  pages={110--121},
  year={2023},
  publisher={Elsevier}
}

@inproceedings{putra2022hate,
  title={Hate speech detection using convolutional neural network algorithm based on image},
  author={Putra, Bagas Prakoso and Irawan, Budhi and Setianingsih, Casi and Rahmadani, Annisa and Imanda, Farradita and Fawwas, Izzu Zantya},
  booktitle={2021 International Seminar on Machine Learning, Optimization, and Data Science (ISMODE)},
  pages={207--212},
  year={2022},
  organization={IEEE}
}

@article{rana2022emotion,
  title={Emotion based hate speech detection using multimodal learning},
  author={Rana, Aneri and Jha, Sonali},
  journal={arXiv preprint arXiv:2202.06218},
  year={2022}
}

@article{mutanga2020hate,
  title={Hate speech detection in twitter using transformer methods},
  author={Mutanga, Raymond T and Naicker, Nalindren and Olugbara, Oludayo O},
  journal={International Journal of Advanced Computer Science and Applications},
  volume={11},
  number={9},
  year={2020},
  publisher={Science and Information (SAI) Organization Limited}
}

@inproceedings{alonso2020hate,
  title={Hate speech detection using transformer ensembles on the hasoc dataset},
  author={Alonso, Pedro and Saini, Rajkumar and Kov{\'a}cs, Gy{\"o}rgy},
  booktitle={International conference on speech and computer},
  pages={13--21},
  year={2020},
  organization={Springer}
}

@inproceedings{ghosh2023transformer,
  title={Transformer-based hate speech detection in assamese},
  author={Ghosh, Koyel and Sonowal, Debarshi and Basumatary, Abhilash and Gogoi, Bidisha and Senapati, Apurbalal},
  booktitle={2023 IEEE Guwahati Subsection Conference (GCON)},
  pages={1--5},
  year={2023},
  organization={IEEE}
}

@article{romim2021hs,
  title={HS-BAN: A Benchmark Dataset of Social Media Comments for Hate Speech Detection in Bangla},
  author={Romim, Nauros and Ahmed, Mosahed and Islam, Md Saiful and Sharma, Arnab Sen and Talukder, Hriteshwar and Amin, Mohammad Ruhul},
  journal={arXiv preprint arXiv:2112.01902},
  year={2021}
}

@inproceedings{ishmam2019hateful,
  title={Hateful speech detection in public facebook pages for the bengali language},
  author={Ishmam, Alvi Md and Sharmin, Sadia},
  booktitle={2019 18th IEEE international conference on machine learning and applications (ICMLA)},
  pages={555--560},
  year={2019},
  organization={IEEE}
}

@article{alam2020bangla,
  title={Bangla text classification using transformers},
  author={Alam, Tanvirul and Khan, Akib and Alam, Firoj},
  journal={arXiv preprint arXiv:2011.04446},
  year={2020}
}

@article{das2022hate,
  title={Hate Speech and Offensive Language Detection in Bengali},
  author={Das, Mithun and Banerjee, Somnath and Saha, Punyajoy and Mukherjee, Animesh},
  journal={arXiv preprint arXiv:2210.03479},
  year={2022}
}

@inproceedings{karim2022multimodal,
  title={Multimodal hate speech detection from Bengali memes and texts},
  author={Karim, Md Rezaul and Dey, Sumon Kanti and Islam, Tanhim and Shajalal, Md and Chakravarthi, Bharathi Raja},
  booktitle={International Conference on Speech and Language Technologies for Low-resource Languages},
  pages={293--308},
  year={2022},
  organization={Springer}
}

@inproceedings{romim2021hate,
  title={Hate speech detection in the Bengali language: A dataset and its baseline evaluation},
  author={Romim, Nauros and Ahmed, Mosahed and Talukder, Hriteshwar and Saiful Islam, Md},
  booktitle={Proceedings of International Joint Conference on Advances in Computational Intelligence: IJCACI 2020},
  pages={457--468},
  year={2021},
  organization={Springer}
}

@article{gitari2015lexicon,
  title={A lexicon-based approach for hate speech detection},
  author={Gitari, Njagi Dennis and Zuping, Zhang and Damien, Hanyurwimfura and Long, Jun},
  journal={International Journal of Multimedia and Ubiquitous Engineering},
  volume={10},
  number={4},
  pages={215--230},
  year={2015},
  publisher={Sandy Bay}
}

@incollection{cavicchio2024rule, 
  title={Rule-Based Systems for Emotion Detection},
  author={Cavicchio, Federica},
  booktitle={Emotion Detection in Natural Language Processing},
  pages={45--52},
  year={2024},
  publisher={Springer}
}

@article{asghar2017sentence,
  title={Sentence-level emotion detection framework using rule-based classification},
  author={Asghar, Muhammad Zubair and Khan, Aurangzeb and Bibi, Afsana and Kundi, Fazal Masud and Ahmad, Hussain},
  journal={Cognitive Computation},
  volume={9},
  number={6},
  pages={868--894},
  year={2017},
  publisher={Springer}
}

@article{alkomah2022literature,
  title={A literature review of textual hate speech detection methods and datasets},
  author={Alkomah, Fatimah and Ma, Xiaogang},
  journal={Information},
  volume={13},
  number={6},
  pages={273},
  year={2022},
  publisher={MDPI}
}

@article{bender2019benderrule,
  title={The\# benderrule: On naming the languages we study and why it matters},
  author={Bender, Emily},
  journal={The Gradient},
  volume={14},
  pages={34},
  year={2019}
}

@article{sanh2019distilbert,
  title={DistilBERT, a distilled version of BERT: smaller, faster, cheaper and lighter},
  author={Sanh, Victor and Debut, Lysandre and Chaumond, Julien and Wolf, Thomas},
  journal={arXiv preprint arXiv:1910.01108},
  year={2019}
}
